\documentclass{article}

     \PassOptionsToPackage{numbers, compress}{natbib}
\usepackage[preprint]{neurips_2025}
\usepackage[utf8]{inputenc} % allow utf-8 input
\usepackage[T1]{fontenc}    % use 8-bit T1 fonts
\usepackage{hyperref}       % hyperlinks
\usepackage{amsfonts}       % blackboard math symbols
\usepackage{amsmath}
\usepackage{nicefrac}       % compact symbols for 1/2, etc.
\usepackage{microtype}      % microtypography
\usepackage{xcolor}         % colors
\usepackage[colorinlistoftodos]{todonotes} %in pdf comments
\usepackage{enumitem}
\usepackage{scalerel}
\usepackage{soul}
\sethlcolor{green}
\usepackage{algorithmic}
\usepackage{algorithm}
\usepackage{booktabs}
\usepackage{multirow}
\usepackage{array}
\usepackage{mathtools}

\newcommand{\ignore}[1]{}
\usepackage{graphics}
\graphicspath{{/figs}}
\usepackage{tikz}
\usepackage{comment}
\usepackage{subcaption}

\usepackage{wrapfig}
\usepackage{listings}
\usepackage{url}
\newcolumntype{H}{>{\setbox0=\hbox\bgroup\color{white}}c<{\egroup}}

\title{On Distributional Dependent Performance of Classical and Neural Routing Solvers}

\author{%
  Daniela Thyssens\thanks{\texttt{thyssens@ismll.uni-hildesheim.de}} \\
  Information Systems and ML Lab, \\
  University of Hildesheim\\
  Hildesheim, Germany \\
  \And
  Tim Dernedde \\
  Information Systems and ML Lab, \\
  University of Hildesheim\\
  Hildesheim, Germany \\
  \AND
  Wilson Sentanoe \\
  University of Hildesheim\\
  Hildesheim, Germany \\
  \And
  Lars Schmidt-Thieme \\
  Information Systems and ML Lab, \\
  University of Hildesheim\\
  Hildesheim, Germany \\
}

\begin{document}

\maketitle

\begin{abstract}
    Neural Combinatorial Optimization aims to learn to solve a class of 
    combinatorial problems through data-driven methods and notably 
    through employing neural networks by learning the underlying 
    distribution of problem instances. While, so far neural methods
    struggle to outperform highly engineered problem specific meta-heuristics,
    this work explores a novel approach to formulate the distribution of 
    problem instances to learn from and, more importantly, plant a structure
    in the sampled problem instances. In application to routing problems, 
    we generate large problem instances that represent custom base problem 
    instance distributions from which training instances are sampled. The test 
    instances to evaluate the methods on the routing task consist of unseen 
    problems sampled from the underlying large problem instance.
    We evaluate representative NCO methods and specialized Operations Research 
    meta heuristics on this novel task and demonstrate that the performance gap 
    between neural routing solvers and highly specialized meta-heuristics 
    decreases when learning from sub-samples drawn from a fixed base node distribution.
\end{abstract}

\section{Introduction}
\label{intro}
The research field of Neural Combinatorial Optimization (NCO), predominantly in the subfield of routing problems, is concerned with learning a parametrized policy to solve a class of combinatorial optimization problems (COPs), thereby saving development effort as well as finding near-optimal solutions faster than traditional algorithms. To train NCO methods, training samples of a particular problem class are randomly generated according to some underlying distribution. For routing problems, this underlying distribution consists in most cases of uniformly at random sampled coordinate values between 0 and 1 (and, in case of the classical vehicle routing problem (VRP) also of uniformly at random distributed integer demand values between 1 and 10) \cite{bello2016neural}.
%With the two-fold motivation of (i) finding reasonably good solutions faster than traditional approaches and (ii) saving development effort and hand-crafted engineering by learning a parameterized policy to trade off computational complexity and optimality in solving combinatorial optimization (CO) problems , the research field of neural combinatorial optimization (NCO) has developed a veritable zoo of methods to solve the vehicle routing problem (VRP) and its variants. 
%While the field of NCO has made considerable progress, the performance gaps to state-of-the-art Operations Research (OR) meta-heuristics are still persistent \cite{Kwon2020.POMO, Choo2022.Simulationguided, Drakulic2023.BQNCO, Ma2023.Learning}. 
%We argue that the way in which the underlying distribution over instances in routing tasks is currently conceptualized is not only unrealistic but also unfavorable for data-driven methods. 
Although NCO has seen substantial advancements, notable performance gaps remain when compared to state-of-the-art meta-heuristics from Operations Research (OR) \cite{Kwon2020.POMO, Choo2022.Simulationguided, Drakulic2023.BQNCO, Ma2023.Learning}. We contend that the prevailing conceptualization of the underlying instance distributions in routing tasks is not only misaligned with real-world conditions, but also suboptimal for the development and evaluation of data-driven approaches.

The training and test instances concurrently used in the literature constitute instance attributes that are sampled independently from each other and thus make for problem instances that are not directly linked through reoccurring attribute patterns.
While the currently employed data generation scheme presents a non-trivial challenge, we argue that it hinders effective learning and pattern recognition, thereby limiting 
%the extent to which the capabilities of neural methods can be meaningfully demonstrated.
%While the currently employed data generation scheme presents a non-trivial challenge, we argue that it hinders effective learning and pattern recognition, thereby limiting 
%limiting the capacity of neural methods to showcase their strengths.
%preventing neural methods from performing to their full capability.
the ability of neural methods to exhibit their full potential.
%Although the currently applied data generating scheme offers a challenging task in itself, we argue that it obstructs learning and pattern recognition, which makes it difficult to showcase the strengths of neural methods in the field. % but also 
%complicates their development, since it is difficult to compare the learning capabilities on unstructured instances.

%Furthermore, we show that also traditional solvers, while more robust to distributional assumptions overall, are not unaffected by them and our experiments show that with more highly structured distributions, ML methods can match/outperform/close the gap/ become competitive to the traditional methods. 
% aim to close the performance gap to state of the art OR methods by formulating a \hl{novel version} of the learning task for NCO methods. 
%Instead we propose to rethink how the learning task for NCO methods is formulated; 

\begin{figure*}[htb!]
  \centering
  \captionsetup{type=figure, width=.9\linewidth} %\includegraphics[width=\textwidth]
  % width=5.9in, height=3.1in, scale=0.9,  width=0.9\linewidth*2  %trim= {1cm 1cm 1cm 1cm} %, scale=0.9
  \includegraphics[width=\linewidth, height=1.9in, trim= {0.8cm 0.4cm 0.7cm 0.4cm}]{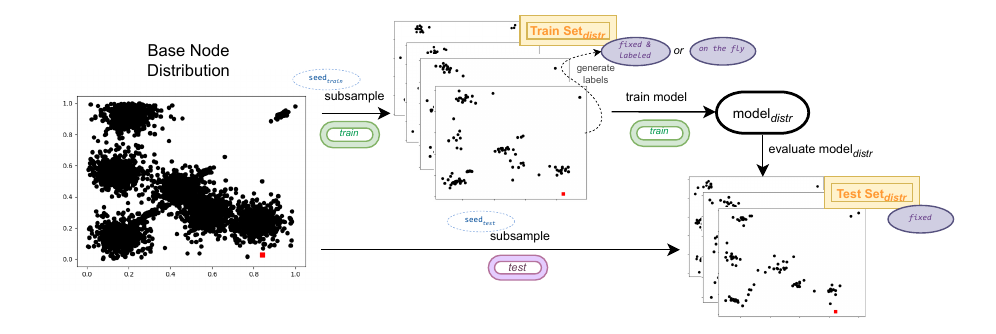}   % of{figure}
  \caption{Overview Subsampling Approach. The Base Node Distribution $\mathcal{G}_{\text{base}}^{\text{X}}$ follows the underlying Random-Clustered (R-C) coordinate distribution introduced in \citet{uchoa2017new}. }
  \label{fig:overview_subsampling}
\end{figure*}

Moreover, we demonstrate that traditional solvers, although generally more robust to distributional assumptions, are nonetheless influenced by them. Our experimental results indicate that, under more structured instance distributions, machine learning-based methods can narrow the gap to — and in some cases even \emph{surpass the performance of} — \emph{classical meta-heuristics}.

Rather than sampling entirely new instances from an underlying distribution, we construct large seed instances — referred to as the \textbf{Base Node Distribution} — based on varying underlying distributions (see Fig. \ref{fig:distributions}). From each seed instance, both training samples and the final test set are obtained via \textit{subsampling}. This setup induces the reocurrence of nodes across instances within a problem class, thereby encouraging NCO methods to internalize and exploit the structural properties of the seed distribution.
%Instead of learning from randomly drawn instances of an underlying distribution, we generate large seed instances, that we denote \textbf{base node distribution}, following different underlying distributions (see Fig. 1), from which training samples as well as the final test set is \textit{subsampled}. This introduces the reoccurrence of nodes that make up the training instances of a problem class, and impels NCO methods to learn the underlying structure of the seed instance. 

From a real-world perspective, the reocurrence of specific nodes reflects many practical scenarios, such as those encountered by logistics service providers who regularly serve established customers, but in varying configurations influenced by seasonal and other periodic trends. For instance, a logistics company may consistently deliver goods to a set of regular retail clients, but the specific delivery schedules, routes, or quantities might vary depending on factors such as holiday shopping seasons, promotional events, or annual sales cycles.

We train three representative NCO methods (POMO \cite{Kwon2020.POMO}, BQ \cite{Drakulic2023.BQNCO}, and NeuOpt \cite{Ma2023.Learning}) on two routing tasks and evaluate them against the two leading state-of-the-art OR algorithms, LKH-3 \cite{helsgaun2017extension} and HGS-CVRP \cite{vidal2022hybrid}, to demonstrate that by reconfiguring the data generation process, ML-based methods can effectively learn the underlying distribution and, in some cases, outperform established OR meta-heuristics.
%to highlight that ML-based methods are capable of effectively learning the underlying distribution of the seed instance, and in certain cases, can surpass the performance of well-established OR meta-heuristics.
%        %providing more significant signals and less noise effects.
        %Besides an already established OR evaluation metric, normalized Primal Integral, the WRAP metric emphasizes the run time efficiency of solvers and captures also the relative improvement 
        %based on the
In summary, we make the following contributions:
\begin{itemize}
  %\item \textit{Novel NCO Learning Task}: We propose to reformulate the learning task for NCO methods to learn from subsamples of a given \textbf{seed instance}, Thus, to learn the problem structure and distribution of subsamples from an underlying seed instance.
    \item \textit{Novel NCO Learning Task}: We reformulate the NCO learning task by having neural methods learn from subsamples of a given \textbf{Base Node Distribution}, enabling them to capture the underlying problem structure and distribution of problem instances through the reocurrence of nodes across instances.
  
    \item \textit{New Datasets}: We introduce new routing problem datasets, providing subsampled training and test sets, along with the data generators used to generate these samples.

    %\item \textit{Evaluation/Insights}: Through a selection of carefully conducted experiments, we demonstrate that the proposed subsampling approach leads to overall improved results with respect to the performance gap to state of the art meta-heuristics and show that they can be outperformed for specific problem distributions.
    \item \textit{Evaluation and Insights}: Through a series of experiments, we show that the proposed subsampling approach narrows the performance gap to state-of-the-art meta-heuristics and, in some cases, even outperforms them for specific problem distributions.
\end{itemize}

\section{Related Work}
\label{sec:related_work}

\paragraph{NCO Data Distributions.}
To effectively apply data-driven methods to combinatorial problems, NCO assumes the existence of an underlying distribution over the instances within a given problem class \cite{Kool2019.Attention, bengioTourdhorizon2021}. This assumption is crucial, as most combinatorial problems are NP-hard, meaning no single solver can be efficient across all possible instances. Even traditional heuristics implicitly rely on assumptions about the instances they address \cite{Yehuda2020.It}. However, for evaluating the capabilities of a learned system, the instance distribution must exhibit sufficient structure to be discoverable \cite{zhou2023towards}.

For VRPs, the distribution is inherently defined by how the following attributes are sampled: (1) the node coordinates, which determine customer locations, (2) the demand from each customer, (3) the position of the depot, and (4) the vehicle capacity. Typically, the coordinates and demands are sampled independently and uniformly at random \cite{Nazari2018.Reinforcement}, while the vehicle capacity is held fixed. This results in a distribution that exhibits minimal structural complexity. While recent works \cite{Xin2021.NeuroLKH, zhou2023towards} have extended their evaluation to more complex distributions, such as Rotation and Explosion \cite{bossek2019evolving} (see Figure \ref{fig:distributions}), all prior works have generated problem instance attributes independently from the underlying distribution. 
%Even alternative distributions, such as Rotation and Explosion (see Figure \ref{fig:distributions}), although not uniform, still sample all attributes and nodes independently.

We propose sampling instances from a base node distribution specific to a problem class, where the base distribution itself follows a designated underlying distribution. This approach complements existing distributions. Our experiments show that sampling from a fixed set of base nodes reduces the performance gap between NCO and classical methods. By introducing structure in a distribution-independent manner, our approach can be \emph{easily integrated into existing evaluation protocols}.
%Instead we propose to sample instances of a problem class from a base node distribution for that problem class, that itself follows a designated underlying distribution. This makes our approach complementary to the mentioned distributions. We show in our experiments that for any of the base distributions, sampling from a fixed set of base nodes decreases the gap between NCO and classical methods. Thus, the approach helps to plant structure in a distribution-independent way, which makes it easy to seamlessly integrate into existing evaluation protocols. 
%The train and test sets used predominantly in the NCO literature consist of \textit{uniformly}-distributed data \cite{bello2016neural, Kool2019.Attention, Kwon2020.POMO, Drakulic2023.BQNCO, Ma2023.Learning}, while some recent works \cite{Xin2021.NeuroLKH, zhou2023towards} extend their evaluation to more complex and real-world distributions, mentioned first in \cite{bossek2019evolving} and \cite{uchoa2017new} respectively. 
%The existing literature concurrently comprises two training approaches; supervised learning Drakulic et al. [10] where a large fixed set of randomly sampled training instances is labeled and reinforcement

\paragraph{Neural Methods for Routing Problems}
NCO methods can roughly be divided into two categories: \textit{improvement}-based methods and \textit{constructive} approaches. Constructive methods create solutions from scratch by sequentially picking nodes with a parameterized policy until a complete solution is constructed \cite{Bresson2021.Transformer, Deudon2018.Learning, Drakulic2023.BQNCO, Jin2023.Pointerformer, Kim2022.SymNCO, kwon2021matrix, Kwon2020.POMO, Luo2023.Neural, Nazari2018.Reinforcement, Sun2023.DIFUSCO, Vinyals2015.Pointer, Xin2021.MultiDecoder, Ye2023.DeepACO}. Different computational trade-offs have been explored. Some approaches compute the entire network at every decision \cite{Drakulic2023.BQNCO, Luo2023.Neural}, some first compute a set of embeddings for all nodes, from which a lighter decoder network computes the decisions, taking into account the updated state via dynamic context features \cite{Berto2024.RouteFinder, Deudon2018.Learning, Jin2023.Pointerformer, Kim2022.SymNCO, Kool2019.Attention, kwon2021matrix, Kwon2020.POMO, Nazari2018.Reinforcement, Xin2021.MultiDecoder} and some methods directly compute a so-called heatmap, encoding the likelihood of each edge belonging to the optimal solution and decode purely from the heatmap without computing the neural network with updated state information again \cite{Kool2022.Deep, Li2023.Distribution, Min2023.Unsupervised, Sun2023.DIFUSCO, Ye2023.DeepACO}.
To increase the accuracy of constructive methods, various searches have been explored, from greedy decoding, sampling, beam search, MCTS and variations thereof \cite{Choo2022.Simulationguided, Kool2019.Attention}. Additionally, hybridizations with heuristic concepts such as Decomposition \cite{Ye2024.GLOP, Zheng2024.UDC}, Dynamic Programming \cite{Kool2022.Deep}, Large Neighborhood Search \cite{Falkner2023.Too, Hottung2022.Neural, Li2021.Learninga} have been made.
Improvement methods on the other hand start off with an initial solution that shall be improved by searching through a neighborhood of that solution. They often rely on existing local search operators. They for instance, learn the acceptance criterion for them, learn a meta controller that picks which operator to use next \cite{falkner2022learning} or directly parameterize the search neighborhood itself, for example, by using predicted edge probabilities and sampling k-opt moves from them \cite{Fu2021.Generalize, Xin2021.NeuroLKH, Ma2023.Learning}.
Both approaches can be trained either via Reinforcement Learning \cite{Kwon2020.POMO, Nazari2018.Reinforcement, Ma2023.Learning} or Imitation Learning from Solver-generated approaches \cite{Fu2021.Generalize, Xin2021.NeuroLKH, Drakulic2023.BQNCO}. 
%For our experiments, we pick popular and state-of-the-art approaches from both categories and evaluate their learning behaviour on our proposed datasets compared to the existing datasets, especially with respect to the relative performance to the classical solvers.
For our experiments, we select popular, state-of-the-art approaches from both categories, improvement-based methods and constructive, and evaluate their learning behavior on our proposed datasets in terms of their relative performance to classical solvers.

\paragraph{Neural Methods and Generalization}
% Most works that aim to improve the generalization capability of trained NCO models focus on the generalization across problem sizes by adapting the model architecture to be scale-conditioned \cite{kim2022scale}, sparsifying attention blocks \cite{bdeir2022attention} or applying curriculum learning \cite{lisicki2020evaluating}. Generalization across distributions is less explored and relies either on knowledge distillation \cite{bi2022learning} and adversarial training \cite{GeislerSSBG22, zhang2022learning} or as recently shown in \citet{zhou2023towards} through applying a meta-learning approach to fine-tune pre-trained models on unseen distributions.
% In this work, we approach generalization from a different angle, in that we propose to train NCO methods to generalize well on unseen instances from the same underlying distribution, while having been trained on a fixed base node distribution. %Moreover we
A significant body of research has focused on enhancing and evaluating the out-of-distribution (OOD) generalization capabilities of learned solvers. Various techniques have been employed, including attention sparsification \cite{bdeir2022attention}, scale conditioning \cite{kim2022scale}, test-time adaptation \cite{Chalumeau2023.Combinatorial, Hottung2022.Efficient, Hottung2024.PolyNet}, meta-learning \cite{zhou2023towards}, adversarial training \cite{GeislerSSBG22, DBLP:conf/aaai/ZhangZW022}, knowledge distillation \cite{bi2022learning}, and curriculum learning \cite{lisicki2020evaluating}. In many cases, OOD behavior is assessed by testing the model on larger instances, where the nodes are sampled according to the underlying distribution of smaller instances. However, this still alters the overall distribution and may change certain problem properties, such as the distance to the closest node, which typically decreases. Some studies also evaluate OOD performance across distributions, but with instances of similar sizes.

In contrast, our evaluation here specifically focuses on in-distribution learning. As discussed earlier, given the NP-hard nature of the problem, it is unrealistic to expect any solver to perform well across all possible instance types. We argue that the primary measure of a model’s performance should be how effectively it fits the instance distribution it is trained on. While robustness to OOD scenarios is undoubtedly important for practical applications, generalization across distributions should always involve a trade-off, particularly given fixed model capacity. We believe that the limited distributions currently employed in the field have, to some extent, hindered model development. In this work, we demonstrate that introducing more structure into the distribution allows learning methods to more effectively exhibit their strengths.

\section{Method}
\label{sec:method}

\subsection{Problem Formulation}
We consider NP-hard binary combinatorial optimization problems, and more specifically routing problems, over graphs. 
% From MOCO paper --> Tim
Let $G=(V,E)$ be a graph, where $V$ are the vertices or nodes and  $E \subseteq V \times V$ are the edges of $G$. The optimization problem consists of a set $\Omega_G \subseteq \{0,1\}^N$ of \emph{feasible solutions} and an \emph{objective function} $f_G \colon \Omega_G \to \mathbb{R}$ and the task is to find an optimal feasible solution $x=(x_1,\dots, x_N) \in \Omega$, by minimizing the objective function:

\begin{equation}
    \min_{x \in \Omega_G} f_G(x).
\end{equation}
For $x \in \{0,1\}^N$, the elements  $x_i$ are also referred to as the binary decision variables.
%Typically, the problem involves the selection of either edges or nodes in the graph leading to either $N=|E|$ or $N=|V|$. 
%routes, $x = \{r_1, ..., r_{|x|}\}$ 
While for the TSP, the task is to select a subset of edges such that the Hamiltonian cycle with minimum distance is found, the task in the VRP is to find a feasible solution that consists of a set of feasible routes $r_k = (x_{k,1}, \ldots, x_{k,R_k})$, which is a sequence of nodes always starting and ending at the depot node, such that $x_{k,1}=0$ and $x_{k,R_k}=0$ and $R_k$ denotes the length of the route. An extended description of the problems can be found in Appendix \ref{app.problem_def}.

\subsection{Subsampling Approach}
\label{sec:subsampling}
% The code and data to reproduce experiments will be available on github\footnote{For now the codebase is available as an anonymized repository: https://github.com/danielathyssens/RA}.

Machine Learning (ML) methods are particularly adept at uncovering patterns and structures inherent in datasets drawn from a specific underlying distribution. This concept underpins the proposed subsampling approach, which differs from existing NCO data sampling methods by subsampling each training (and test) instance from the same distribution of vertices, referred to as the base node distribution, rather than directly from an underlying distribution. The goal of this approach is to enhance the recurrent presence of previously encountered nodes during training, thereby encouraging the model to learn the spatial and relational structure of the base node distribution. 
Furthermore, we highlight that the reocurrence of customer nodes in routing tasks not only strengthens the ability of NCO methods to learn patterns but also reflects the practical requirements of real-world applications.

Formally, we define a base node distribution $\mathcal{G}_{\text{base}}^{j}$ that is sampled from an underlying distribution $\{\mathcal{D}_j\}^{|\mathcal{D}|}_{j=1}$: 
% where k refers to the underlying distribution that the base node distribution follows.
% Let $G\base:= (V\base,E\base)$ be a graph and $S\base:=\{0,1\}^N$
% the solution space of a COP on this graph, i.e.,
% $N = |E\base|$ be the number of edges (for TSP) or
% $N = |V\base|$ the number of vertices (for MIS).
% TODO: CVRP.
% Let $G=(V,E)\subseteq G\base$ a subgraph, i.e., $V\subseteq V\base$
% and $E:= \{ e\in E\base \mid e_1, e_2\in V\}$.
\begin{equation}
    \mathcal{G}_{\text{base}}^{j} \sim \mathcal{D}_{j}
\end{equation}

where $\mathcal{D}_{j}$
represents the coordinate and demand distributions associated with the $N_{\text{base}}$ customer nodes $\{V_b\}^{N_{\text{base}}}_{b=0}$ 
sampled for the base node distribution. In the case of the CVRP, node $V_0$ in $\mathcal{G}_{\text{base}}^{j}$ is the designated depot node which remains fixed for subsampled CVRP problem instances.

Given the base node distribution $\mathcal{G}_{\text{base}}^{j}$ we sample nodes through $f_{\text{sample}}$ to form a COP instance $G_i$ of problem-size $N$:

\begin{equation}
\begin{aligned}
    G_i^{j} = V_1, V_2, \dots, V_{N} & \sim f_{\text{sample}}(\mathcal{G}_{\text{base}}^{j}) \quad \text{for } i = 1, \dots, L\\
    %X_1, X_2, \dots, X_{L_{N}} & \sim f_{\text{sample}}(\mathcal{G}_{\text{nodes}}^{1,...,n-1}) \\
    %\mathbf{G}^{\text{\tiny{te}}}_{\text{nodes}} \sim f_{\text{sample}}(1, n-1) 
\end{aligned}
\end{equation}
where $L$ is the sample size or dataset length and $f_{\text{sample}}$ corresponds to the uniform sampling function, such that $G_i^{j} \stackrel{i.i.d.}{\sim} \mathcal{G}_{\text{base}}^{j}$. We note that for the CVRP, the depot node $V_0$ is always added to the subsampled node set $G_i^{j}$.
Finally we define the train and test sets for an underlying distribution $\mathcal{D}_j$ and a respective base node distribution $\mathcal{G}_{\text{base}}^{j}$ as follows:

\begin{equation}
\begin{aligned}
    \mathbf{G}_{\text{\tiny{train}}}^{j} := \{G_i^{j}\}_{i=1}^{L_{\text{\tiny{train}}}}, \quad
    %\mathbf{G}_{\text{\tiny{test}}}^{j} := \{G_i^{j}\}_{i=1}^{L_{\text{\tiny{test}}}}\\
    %\quad
    \mathbf{G}_{\text{\tiny{test}}}^{j} := (G^j_i)_{i=L_{\text{\tiny{train}}} + 1}^{L_{\text{\tiny{train}}}+L_{\text{\tiny{test}}}}
\end{aligned}
\end{equation}
%\todo{other variable name than X - already used as feasible solutions..}
%$ f_{\text{sample}} = \mathcal{U}$.
\vspace{0.1cm}

\begin{figure*}[ht!]
  \centering
  \captionsetup{type=figure, width=.9\linewidth} %\includegraphics[width=\textwidth]
  % width=5.9in, height=3.1in, scale=0.9,  width=0.9\linewidth*2  %trim= {1cm 1cm 1cm 1cm} %, scale=0.9
  \includegraphics[width=\linewidth, trim= {0.2cm 0.01cm 0.1cm 0.1cm}]{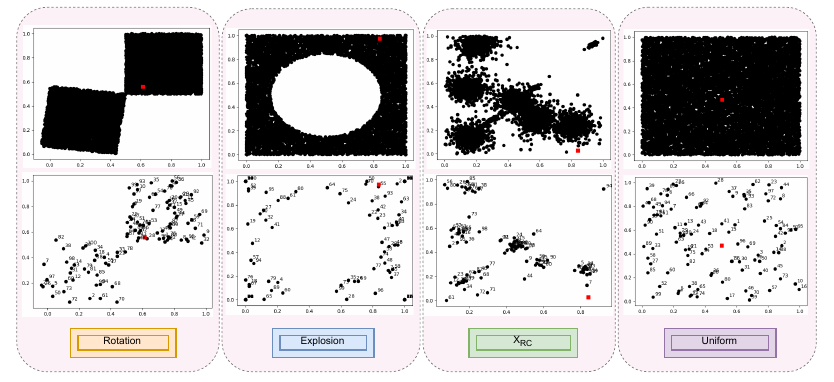}   % of{figure}
  \caption{Base Node Distributions $\mathcal{G}_{\text{base}}^{j}$ (top) from which routing problems $G_i$ are subsampled (bottom).}
  \label{fig:distributions}
\end{figure*}

NCO methods have employed various sampling strategies for training on routing tasks. Notably, RL-based approaches are primarily trained on data generated during each epoch of training \cite{Nazari2018.Reinforcement, Kool2019.Attention, Ma2023.Learning}. To efficiently incorporate subsampling into the training of different NCO methods, we define a generic NCO Dataset class, which includes data generators and samplers to facilitate subsampling within the RL training process.
%if applicable, transforms the subsampled instances $\mathbf{G}_{\text{\tiny{epoch}}}^{j}=\{G_i\}^{L_{\text{epoch}}}_{i=1}$ for each training epoch.
Algorithm \ref{algo:subsample} illustrates how the subsampling approach is integrated into the RL training pipeline. For NCO methods that rely on supervised learning, we generate fixed training datasets consisting of subsampled instances $\{G_i^{j}\}_{i=1}^{L_{\text{\tiny{train}}}}$ and their corresponding labels $\{Y_i^{j}\}_{i=1}^{L_{\text{\tiny{train}}}}$ that are generated with LKH3 \cite{helsgaun2017extension} for the TSP and HGS-CVRP \cite{vidal2022hybrid} for the CVRP.

\begin{algorithm}
    \caption{Subsampling for RL Training}\label{algo:subsample}
    \begin{algorithmic}
        \STATE {\bfseries Input:} Problem Class $P$, Underlying distribution $\mathcal{D}_{j}$ %,Base Node Distribution $\mathcal{G}_{\text{base}}^{j}$, Batch Size $b$
        \vspace{2mm}
        \STATE $\text{Dataset}_{P,j} \gets \text{DatasetClass}(P, j)$ \COMMENT{$\textit{init } \text{Dataset with } P \text{ and } j$}
        \STATE $\mathcal{G}_{\text{base}}^{j} \gets \text{Dataset}_{P,j}\text{.Base\_Node\_Distribution}$;
        %\STATE $\ \gets \text{Dataset}_{P,j}\text{.Base\_Node\_Distribution}$;
        \STATE $\theta^{(0)} \gets $
        \COMMENT{\textit{initialize model}}

        \FOR{$\text{epoch}\gets0$ to $\text{number\_epochs}$}
        \STATE $\mathbf{G}_{\text{\tiny{epoch}}}^{j} \gets f_{\text{sample}}(\mathcal{G}_{\text{base}}^{j})$ \COMMENT{\textit{sub-sample a $L_{\text{epoch}}$ train instances}}
        \STATE $\theta^{(\text{epoch})} \gets \text{train\_epoch}(\mathbf{G}_{\text{\tiny{epoch}}}^{j})$
        \ENDFOR

        \STATE {\bfseries Return:} $\theta^{(\text{epoch})}$
    \end{algorithmic}
\end{algorithm}

Figure \ref{fig:overview_subsampling} illustrates an overview of the subsampling approach for an exemplary base node distribution of $N_{\text{base}}=10000$ nodes. The base node distribution $\mathcal{G}_{\text{base}}^{\text{X}}$ in Fig. \ref{fig:overview_subsampling} is sampled from an underlying distribution that mimics real-world customer distributions described in \citet{uchoa2017new} and specifically involves clusters of customer locations. The train and test instances are subsampled with different seeds respectively to avoid training on test. The train set is then either subsampled on the fly for RL approaches as described in Algorithm \ref{algo:subsample} or collected and labeled apriori for supervised learning approaches. After a model is trained on subsamples of the base node distribution it is tested on a fixed test set $\mathbf{G}_{\text{\tiny{test}}}^{\text{X}}$. In the subsequent experiments section, the notation of subsampled test sets is extended to incorporate different sizes, $N_{\text{base}}$, of the base node distributions, such that $\mathbf{G}_{\text{\tiny{200}}}^{\text{X}}$ represents the test set subsampled from a base node distribution $\mathcal{G}_{\text{base}}^{\text{X}}$ with $N_{\text{base}}=200$.

\section{Experiments}
\label{experiments}
To assess the effectiveness of the proposed subsampling approach for NCO, we sample Base Node Distributions from four distinct distributions $\mathcal{D}_j$, as illustrated in Figure \ref{fig:distributions}. We evaluate four representative NCO models on the TSP and CVRP: two established constructive methods, POMO \cite{Kwon2020.POMO} and BQ \cite{Drakulic2023.BQNCO}, and two search methods, SGBS-EAS \cite{Choo2022.Simulationguided} and NeuOpt \cite{Ma2023.Learning}. Additionally, we benchmark the subsample-trained NCO methods against the state-of-the-art OR solvers, LKH3 \cite{helsgaun2017extension} and HGS-CVRP \cite{vidal2022hybrid}.

\subsection{Experiment Setup}

\paragraph{\textbf{Distributions}}
We evaluate our proposed approach on four diverse underlying distributions (Fig. \ref{fig:distributions}) that are commonly used in the literature, such that $\{\mathcal{D}_j\}^{4}_{j=1}$ = \{Rotation, Explosion, X, Uniform\}.
The \emph{Rotation} location distribution was first introduced in \citet{bossek2019evolving} to diversify the pool of node distributions for the TSP. Following \citet{zhou2023towards} we introduce randomly sampled demands to be associated with the coordinate nodes for the CVRP. Specifically we sample customer demands from a discrete uniform distribution \{1, \ldots , 9\}.
Similarly, the \emph{Explosion} coordinate distribution, also introduced in \citet{bossek2019evolving} is employed for the TSP and complemented with randomly sampled integer demands for the CVRP.
We incorporate two variations of the \emph{X} distributions, as proposed in \citet{uchoa2017new}, to better approximate real-world scenarios. For the TSP, we use Clustered customer coordinates ($X_{\text{C}}$) and for the CVRP we generate a base node distribution with a randomly selected depot location (R), half-Random, half-Clustered node coordinates (depicted as $X_{\text{RC}}$ in Fig. \ref{fig:distributions}) and unitary customer demands ($X_{\text{RRC0}}$).
The \emph{Uniform} distribution is the most conventional distribution used to generate routing problems in the field of NCO. Following \citet{Nazari2018.Reinforcement} and subsequent works (\cite{Kool2019.Attention, Kwon2020.POMO,Ma2023.Learning}) we sample node coordinates uniformly in the unit square $\mathcal{U}(0,1)$ for the TSP and add uniformly sampled integer demands \{1, \ldots , 9\} as node features for the CVRP.
Further details on the underlying distributions are discussed in Appendix \ref{app:data_generation}.

\renewcommand{\arraystretch}{1.5}
\begin{table}[htb]
  \centering
  \tiny
  \caption{Datasets used in the Experiments for NCO-Subsampling. In total we are testing the methods on 14 subsampled testsets. $N$ refers to the graph size of the subsampled problems that make up a test set. $N_{\text{base}}$ refers to the size of the the Base Node Distribution (number of base nodes) the test sets are subsample from. We note that the size of the test sets is coherently set to 128 instances.}
  %\resizebox{\columnwidth*2}{!}{
  \resizebox{0.65\columnwidth}{!}{
    \begin{tabular}{l|cccl}
        Dataset                        & $N$ & $N_{\text{base}}$ & Problem Class & Distribution\\ %& \# Test Instances \\   %& targets   \\
      \midrule
      %$\mathbf{G}_{\text{\tiny{10k}}}^{\text{\footnotesize{Unif}}}$  &  Uniform \cite{Kool2019.Attention}    & 100          & 10000 & \\  % & 128 \\ % & \greencheck    \\
      \textbf{$\mathbf{G}_{\text{{10k}}}^{\text{Unif}}$}    & 100          & 10000 & TSP, CVRP &  Uniform \cite{Kool2019.Attention}\\  % & 128 \\ % & \greencheck    \\
      \textbf{$\mathbf{G}_{\text{{10k}}}^{\text{X}}$}       & 100          & 10000 & TSP, CVRP  &  $\text{X}_{\text{C}}$, $\text{X}_{\text{RRC0}}$   \cite{uchoa2017new}\\  % & 128 \\ % & \greencheck   \\
      \textbf{$\mathbf{G}_{\text{{200}}}^{\text{X}}$}       & 100          & 200   & TSP, CVRP &  $\text{X}_{\text{C}}$, $\text{X}_{\text{RRC0}}$ \cite{uchoa2017new}\\ 
      \textbf{$\mathbf{G}_{\text{{10k}}}^{\text{Exp}}$}     & 100          & 10000 & TSP, CVRP &  Explosion \cite{bossek2019evolving}\\  % & 128 \\ % & \greencheck   \\
      \textbf{$\mathbf{G}_{\text{{200}}}^{\text{Exp}}$}     & 100          & 200 & TSP, CVRP &  Explosion \cite{bossek2019evolving}\\  % & 128 \\ % & \greencheck   \\
      \textbf{$\mathbf{G}_{\text{{10k}}}^{\text{Ro}}$}      &   100        & 10000 & TSP, CVRP  &   Rotation \cite{bossek2019evolving}\\ %  & 128 \\ % & \greencheck   \\
            \textbf{$\mathbf{G}_{\text{{500}}}^{\text{Ro}}$}     &   100   & 500 & TSP, CVRP  &   Rotation \cite{bossek2019evolving}\\ %  & 128 \\ % & \greencheck   \\
    \end{tabular}}

  \label{tab:datasets}
%\end{table*}
\end{table}%

\paragraph{\textbf{Baselines}}
We select four representative NCO baselines for the experiments and retrain them according to the subsampling approach for each of the base node distributions described above.
As \emph{construction-based approaches} we implement subsampling for POMO \cite{Kwon2020.POMO} and BQ \cite{Drakulic2023.BQNCO}.
Policy Optimization with Multiple Optima (POMO) \cite{Kwon2020.POMO} improves the Attention Model (AM) \cite{Kool2019.Attention} by introducing a new RL-based training and inference mechanism. The baseline function in the policy gradient is adjusted, such that it is averaging over multiple rollouts with different starting nodes for a problem instance to get a better baseline estimate. 
%POMO still constitutes a strong constructive baseline that leverages the solution symmetry of the optimal solution notably for the TSP. 
As supervised learning-based approach, BQ \cite{Drakulic2023.BQNCO} rethinks the encoder-decoder paradigm of earlier approaches by computing the entire problem (graph $G_i$) at every construction step with a deep neural network, which updates the input state according to the current construction step, which allows the model to solve a sequence of subproblems during the construction process.
Notably, we generate targets for each of the training sets subsampled from the base node distributions to train BQ, which we will make available in the supplementary material.
Concerning \emph{search and improvement approaches} we select SGBS \cite{Choo2022.Simulationguided} and NeuOpt \cite{Ma2023.Learning} as neural baselines for the subsampling approach.
Simulation Guided Beam Search (SGBS) improves pretrained constructive models during inference by evaluating and pruning candidate solutions of a post-hoc beam search. Combined with Efficient Active Search (EAS) \cite{eas}, which serves as an additional fine-tuning strategy, SGBS-EAS forms an efficient and high-performing search approach for neural-based methods and works out of the box for trained POMO models.
NeuOpt \cite{Ma2023.Learning} improves initial solutions iteratively by parametrizing a local search operator (k-opt) through a neural network. Moreover, NeuOpt allows for temporary constraint violations and thereby the generation of infeasible solutions, extending the search space of the neural improvement method. The implementation details and notably the model-specification are explained in Appendix \ref{app:baselines}.
As representative OR heuristics, we include the well-known meta-heuristic LKH3 \cite{helsgaun2017extension} and, specifically for CVRP, Hybrid-Genetic-Search for the CVRP (HGS-CVRP) \cite{vidal2022hybrid}.
Table \ref{tab:baselines} summarizes the baselines and their respective properties.

\begin{table}[ht!]
  \centering
  \caption{Baselines used in the Experiments for NCO-Subsampling. "C" refers to construction-based approaches and "LS" to local search approaches. The training procedures are categorized into Reinforcement Learning (RL) and Supervised Learning (SL).  }
  \resizebox{0.7\columnwidth}{!}{
    \begin{tabular}{l|cccc}
      Baseline                           & Area & Type & Train Type & Model Variant \\
      \midrule
      POMO \cite{Kwon2020.POMO}           & ML   & C    & RL         & augmentation (x8)   \\
      BQ \cite{Drakulic2023.BQNCO}        & ML   & C    & SL         & beam (16)      \\
      SGBS+EAS \cite{Choo2022.Simulationguided} & ML   & LS & RL         & augmentation (x8) \& 28 EAS loops \\
      NeuOpt \cite{Ma2023.Learning}         & ML   & LS   & RL         & $T_{\text{MAX}}$=5000\\
      LKH3 \cite{helsgaun2017extension}  & OR   & LS   & -          & - \\
      HGS \cite{vidal2022hybrid}         & OR   & LS   & -          & - \\
    \end{tabular}}
  \label{tab:baselines}
\end{table}%

\paragraph{\textbf{Metrics \& Hardware}}
%the Primal Integral (PI) \cite{berthold2013measuring} value
The performance of the routing solvers is measured by the average per-instance percentage gap of objective costs. The gap is computed relative to the objective cost provided by the respective specialized meta-heuristic for the given problem class; LKH3 \cite{helsgaun2017extension} for the TSP and HGS \cite{vidal2022hybrid} for the CVRP. 
Since the experiments benchmark search-based methods that operate on the basis of a specified search time limit, we set pre-defined per-instance runtime budgets ($T_{\scaleto{\text{MAX}}{3.5pt}}$) for each experimental run that are coherent across OR and ML methods. We note that, to compensate constructive methods for not utilizing time budget, we add a simplistic search heuristic for constructive methods, in the form of Simulated Annealing.
%For the benchmarking experiments against OR metaheuristic solvers (section \ref{exp:vs_or}), we additionally document \hl{the fraction of better-found solutions found by NCO methods} \todo{name for this?}.
%For the objective cost, we record the \textit{average} solution cost over the respective test dataset. The gap is computed relative to the best solution found during the course of the experiments.
Inference runs are performed on a single Nvidia RTX 3060 GPU and 13th Gen Intel(R) Core(TM) i7-1355U CPU. 
%Training for the baseline models are performed on a single Nvidia RTX 3090 GPU and AMD EPYC 7543P CPU or on a node with 8 Nvidia A4000 16GB GPUs.
%\paragraph{}

% Results are averaged over 3 runs for a time budget of $T_{\scaleto{\text{MAX}}{3.5pt}} = 2.4N$ (s).
%while \textbf{$\{G_i\} \sim \mathcal{D}_\text{j}$} refers to the test set sampled from $\mathcal{D}_\text{j}$ directly

\subsection{Experimental Results: Subsampling Versus Full Distribution Training}
\label{exp:subsampling}

This subsection presents experimental results evaluating the effectiveness of subsampling from base node distributions, in contrast to training directly on the underlying distribution. 
Table \ref{table:sub_vs_model} presents a comparative evaluation 
%of models trained on subsampled base node distributions ($\text{method}{\text{sub}}$) versus those trained on the full underlying distribution ($\text{method}$).
where each neural baseline method is showcased in two training configurations—$\text{method}_{\text{sub}}$, trained on subsampled data, and $\text{method}$, trained on the full underlying distribution. Both variants are evaluated on test sets drawn from the base node distributions $\mathbf{G}_{N_{\text{base}}}^{j}$ for $j \in \{\text{Rotation}, \text{Explosion}, \text{X}, \text{Uniform} \}$ respectively. 

\begin{table*}[htb!]  %[b]{0.5\textwidth}
  \caption{Comparison of the subsampling approach to conventional data sampling in training. \textbf{$\mathbf{G}_{N_{\text{{base}}}}^{j}$} where $j \in \{\text{Uniform, Rotation, Explosion, X}\}$ refers to the test set sampled from the base node distribution. The runtime limit for all models is set to $T_{\scaleto{\text{MAX}}{3.5pt}} = 10$ seconds per instance. The Relative Gap (in $\%$) is computed to LKH and HGS, respectively for the TSP and CVRP. Smaller values are better. \textbf{Best} performance, \textit{best} model variant. }
  \centering
  \resizebox{0.95\columnwidth}{!}{
    \scriptsize
    \begin{tabular}{lll|llll|llllH}  
      \toprule
      % & \textbf{Gap (\%)}    & \multicolumn{4}{c|}{\% Gap} %& \multicolumn{4}{c}{Times}                          \\
      & & \textbf{Gap (\%)}     & \multicolumn{4}{c|}{TSP}   & \multicolumn{5}{c}{CVRP} \\ %&    &          & \\
      \cmidrule(lr){4-7} \cmidrule(lr){7-12} %\midrule
      & & Dataset            & \textbf{$\mathbf{G}_{\text{{10k}}}^{\text{Unif}}$}      & \textbf{$\mathbf{G}_{\text{{10k}}}^{\text{Ro}}$}    & \textbf{$\mathbf{G}_{\text{{10k}}}^{\text{Exp}}$}  & \textbf{$\mathbf{G}_{\text{{10k}}}^{\text{X}}$} & \textbf{$\mathbf{G}_{\text{10k}}^{\text{Unif}}$}      & \textbf{$\mathbf{G}_{\text{{10k}}}^{\text{Ro}}$}    & \textbf{$\mathbf{G}_{\text{{10k}}}^{\text{Exp}}$}  & \textbf{$\mathbf{G}_{\text{{200}}}^{\text{X}}$}  & \textbf{$\mathbf{G}_{\text{{10k}}}^{\text{X}}$}    \\      % \\ & POMO         & BQ        & SGBS      & NeuOpt   \\   %& CW   & LKH     & HGS       \\
      %\parbox[t]{1mm}{\multirow{13}{*}{\rotatebox[origin=c]{90}{\textbf{TSP}}}}  \\
      \midrule
      &\parbox[t]{1.5mm}{\multirow{2}{*}{\rotatebox[origin=c]{90}{\phantom{nn}}}}  &    &     &      &      &     &     &      &      &    \\
      && $\text{POMO}_{\text{sub}}$ &  1.2190         &   \textit{1.646}   &  \textit{1.675}    &  4.598    
                             &  5.848          &   \textit{4.559}   &  \textit{2.702}    &  1.354            & -2.229 \\
      %&&  \textbf{\{G_i\} \sim $\mathcal{D}_\text{Unif}$}    &  2.615   48}\\
      && POMO                &  \textit{0.144} &   1.673            &  3.326             &  \textit{3.394}                           
                             &  \textit{5.363} &   4.608            &  2.827             &  \textbf{\textit{0.449}}   & \textit{-4.060} \\
      &\parbox[t]{1.5mm}{\multirow{2}{*}{\rotatebox[origin=c]{90}{\phantom{nn}}}}  &    &     &      &      &     &     &      &      &    \\ 
      && $\text{SGBS}_{\text{sub}}$ & 0.9280          &   \textit{0.288}   &  \textit{0.761}    & 2.517                                
                             & 5.206           &   \textit{3.045}   &  \textit{2.278}    & \textit{1.843}    &   -3.041\\
      && SGBS                & \textit{\textbf{0.0544}} &   0.487   &  1.474             & \textit{1.331}         
                             & \textit{4.143}  &   4.006            &   2.514            & 9.461             &  \textbf{-4.265}\\     
      &\parbox[t]{1.5mm}{\multirow{2}{*}{\rotatebox[origin=c]{90}{\phantom{nn}}}}  &    &     &      &      &     &     &      &      &    \\ %  &      &      \\
      && $\text{BQ}_{\text{sub}}$   &   0.111          &  0.103             &  \textbf{0.139}    &   \textbf{0.099}                         
                             &   \textbf{1.675} &  \textbf{1.440}    &  2.996             &   1.237         &   3.093\\
      %&& BQ    &     &    &      &           \\
      &\parbox[t]{1.5mm}{\multirow{2}{*}{\rotatebox[origin=c]{90}{\phantom{nn}}}}  &    &     &      &      &     &     &      &      &    \\ %  &      &      \\
      && $\text{NeuOpt}_{\text{sub}}$ &  \textit{0.199} &   \textbf{\textit{0.074}}   &  \textit{0.536}   &   2.355      
                               & 6.546           &   \textit{1.878}            &   7.237           &  10.027       &   -0.782\\
      && NeuOpt                &  0.302          &    0.192                    &  0.549            &   \text{0.323}         
                               & \textit{1.895}  &   2.043                     &  \textbf{\textit{1.636}}  & 2.933 &  \textit{-3.368}\\
      \cmidrule(lr){2-12} %\midrule
      %& \parbox[t]{1.5mm}{\multirow{6}{*}{\rotatebox[origin=c]{90}{\textbf{\{G_i\} \sim $\mathcal{D}_\text{j}$}}}}  &      &     &      &      &     &     &      &      &     \\ %  &          \\

      %\bottomrule
      %&& AVG                 &     &      &      &   &     &      &      & \\
    \end{tabular}}
  %\captionsetup{type=table, width=.9\linewidth}  %of{table}
  \label{table:sub_vs_model}
  \end{table*}

Across a range of neural solvers and problem distributions, the goal is to assess whether focusing on structured subsamples from $\mathbf{G}_{\text{base}}^j$ leads to more efficient and (in-distribution) generalizable training. We note that due to the SL training scheme of BQ, which necessitates the generation of labels, we only trained the model through subsampling for the respective distributions.
Overall, we observe that subsampled variants often yield competitive—and often superior—performance compared to their fully trained counterparts. Notably, in the TSP setting, methods such as SGBS and NeuOpt exhibit marked improvements in several tasks when trained on subsampled distributions, suggesting that subsampling can lead to sharper optimization and better alignment with the test distribution. For the CVRP, SGBS\textsubscript{sub} achieves a lower relative gap than its full-distribution version in three out of four settings. Similarly, NeuOpt\textsubscript{sub} attains the best result on the CVRP Rotation dataset.
However, on certain CVRP ands TSP benchmarks, models trained on the full distribution perform better—likely due to the increased complexity and diversity of instances where broad exposure to the full distribution aids generalization and the focus on reoccurring structural relations between nodes does not improve pattern recognition, as is the case for the Uniform distribution. In general, it is expected that the impact on uniformly distributed instances is less pronounced, as there is no direct and inherent structure to learn. Notably also the general level of the CVRP performance gaps for the \textbf{$\mathbf{G}_{\text{{10k}}}^{\text{Unif}}$} is higher.
Nonetheless, even in those cases, the performance gap is often small, and BQ\textsubscript{sub} and NeuOpt\textsubscript{sub} still produce strong results on multiple CVRP benchmarks.
Taken together, these findings indicate that subsampling from structured base node distributions is a promising training strategy. It can lead to more efficient learning, reduced variance, and in several settings, even improved test performance—all without the need to model the full complexity of the underlying data distribution, particularly in settings where computational efficiency or domain-specific structure can be leveraged.
% The results for the TSP clearly indicate the usefulness of the proposed sampling approach for most underlying distributions. While the performance of sub-model variants on the Explosion and Rotation distributions demonstrate significant lifts, notably highlighting $\text{NeuOpt}_{\text{sub}}$ for the Rotation distributed test instances where the gap can be reduced to 0.074, the performances across models for the more challenging clustered distribution ($\text{X}_{\text{C}}$) and the unstructured uniform distribution are mixed.
%Concerning the CVRP, the right side of table \ref{table:sub_vs_model} depicts a similar pattern; For the highly structured rotation and explosion distributions the subsampling-trained model variants excel, while the subsampling impact is less pronounced for the $\text{X}_{\text{RRC0}}$ and the uniform distribution.

\subsection{Experimental Results: Comparison to Classical solvers}
\label{exp:vs_or}
The results in Table \ref{table:sub_vs_OR} compare the performance of ML-based models trained on base node distributions against established OR metaheuristics (LKH for TSP and HGS for CVRP), 
%which are known to obtained best performances on the TSP and CVRP respectively, 
under three increasing time budgets. The relative gap (in \%) to these strong baselines is reported across various benchmark distributions and problem sizes.
%mark a significant milestone for neural combinatorial optimization: ML-based methods not only close the gap to classical OR metaheuristics but, in select settings, outperform them. This is reflected in the negative relative gaps, which indicate better solutions (lower costs) than those produced by LKH (for TSP) or HGS (for CVRP) under a shared runtime budget of 10 seconds per instance.
The main result from Table \ref{table:gap_cvrp} is that ML-based solvers can not only often match but also surpass classical optimization methods in several configurations, demonstrated by negative relative gaps, where ML solutions outperform the OR baselines in objective value within the same runtime budget.
This trend is most pronounced for the CVRP:
\begin{itemize}
    \item     On X\textsubscript{10k}, POMO, SGBS, and NeuOpt consistently achieve negative gaps across all three time budgets. Notably, SGBS reaches $-4.22$\% under the 50s runtime budget, marking a substantial improvement over HGS.
    \item     Even under a tight 0.7s budget, BQ achieves a slight negative gap on Uchoa\textsubscript{200}, while POMO and SGBS also outperform classical solvers at higher budgets on this distribution.
\end{itemize}

%Concerning the TSP, ML models do not yet achieve negative gaps, BQ\textsubscript{sub} closes the gap to nearly zero, achieving $−0.01$\% on X\textsubscript{200} under a $0.7$s time limit — demonstrating remarkable efficiency given the tight computational constraint.

Concerning the TSP, ML models do not yet achieve negative gaps; BQ\textsubscript{sub} closes the gap to nearly zero, achieving $-0.01$\% on the \textbf{$\mathbf{G}_{\text{{200}}}^{\text{X}}$} test set under a $0.7$\,s time limit - demonstrating remarkable efficiency given the tight runtime constraint.
Table~\ref{table:sub_vs_OR} shows that when trained on structured distributions, learned policies can surpass even specialized OR heuristics, particularly where data-driven models capture nuanced structural priors. At the same time, the enduring strength of classical solvers reflects their maturity and general robustness, especially across diverse distributions or extended runtimes. Still, the ability of neural models to match - or occasionally outperform - these baselines in constrained, aligned settings suggests a strong complementary role, especially in domain-specific or real-time applications.
While traditional methods retain an edge in average-case performance, ML solvers emerge as viable, adaptive alternatives for targeted deployment.
% On one hand, the results in Table \ref{table:sub_vs_OR} suggest that learned policies, when trained on structured node distributions, can outperform highly-specialized OR heuristics in solution quality, notably in domains where learned representations capture nuanced problem structures. On the other hand, they also reflects the well-established robustness and maturity of classical solvers. In many settings, especially under broader distributional coverage or longer time budgets, these methods remain highly effective. Nonetheless, the fact that learned models can occasionally match or even surpass them under constrained, distribution-aligned conditions highlights a promising complementary role for neural methods—particularly in specialized or real-time deployment scenarios.

%\textbf{$\mathbf{G}_{\text{{200}}}^{\text{X}}$}
\begin{table}[ht!]  %[b]{0.5\textwidth}
  \caption{Comparison of ML-based models trained on base node distributions with classical OR metaheuristics. Negative gaps (bolded) indicate cases where learned models outperform LKH (TSP) or HGS (CVRP) under the same runtime budget. The subscript of the respective Dataset refers to the size of the base node distribution ($N_{\text{base}}$).}
   \label{table:sub_vs_OR}
  \begin{subtable}{.461\linewidth}
  \renewcommand{\arraystretch}{1.3}
  \centering
  \caption{TSP}
    \label{table:gap_tsp}
  \resizebox{\columnwidth}{!}{
  %\scalebox{0.8}{
    %\renewcommand{\arraystretch}{1} % Default value: 1
%    % \tiny
    %\small
    % \footnotesize
    \setlength\tabcolsep{3pt}
    \begin{tabular}{ll|llll|l}  %lllllllll
      \toprule
      & \textbf{Gap (\%)}    & \multicolumn{4}{c|}{ML} & \multicolumn{1}{c}{OR}                             \\
      & Dataset                 & POMO\textsubscript{sub}   & BQ\textsubscript{sub}                & SGBS\textsubscript{sub}       & NeuOpt\textsubscript{sub}    & LKH       \\
      \midrule
      \parbox[t]{2mm}{\multirow{8}{*}{\rotatebox[origin=c]{90}{\textbf{$T_{\scaleto{\text{MAX}}{3.5pt}} = 0.7$}}}}  &    &     &   &   &   &     \\
      & \textbf{$\mathbf{G}_{\text{{200}}}^{\text{Exp}}$}  & 4.322   & 3.116            & 2.941      & 7.588     & 0.0    \\
      & \textbf{$\mathbf{G}_{\text{{10k}}}^{\text{Exp}}$}  & 1.022   & 0.152            & 1.671      & 0.621     & 0.0    \\
      & \textbf{$\mathbf{G}_{\text{{200}}}^{\text{X}}$}      & 1.142   & \textbf{-0.0100} & 7.394      & 15.827    & 0.0 \\
      & \textbf{$\mathbf{G}_{\text{{10k}}}^{\text{X}}$}     & 4.588   & 0.104            & 4.525      & 3.549     & 0.0   \\
      & \textbf{$\mathbf{G}_{\text{{500}}}^{\text{Ro}}$}   & 2.987   & 1.621            & 1.800      & 7.214     & 0.0   \\
      & \textbf{$\mathbf{G}_{\text{{10k}}}^{\text{Ro}}$}   & 1.645   & 0.100            & 1.561      & 0.451     & 0.0 \\
      & \textbf{$\mathbf{G}_{\text{{10k}}}^{\text{Unif}}$}   & 2.680   & 0.147            & 2.680      & 0.369     & 0.0   \\
      \parbox[t]{2mm}{\multirow{8}{*}{\rotatebox[origin=c]{90}{\textbf{$T_{\scaleto{\text{MAX}}{3.5pt}} = 5$}}}}  &    &     &   &   &   &     \\
      & \textbf{$\mathbf{G}_{\text{{200}}}^{\text{Exp}}$}  & 4.196   &  2.016           & 1.569      &  0.737   & 0.0   \\
      & \textbf{$\mathbf{G}_{\text{{10k}}}^{\text{Exp}}$}  & 1.022   &  0.139           & 0.931      &  0.536   & 0.0   \\
      & \textbf{$\mathbf{G}_{\text{{200}}}^{\text{X}}$}      & 1.193   &  0.046           & 6.321      & 13.19    & 0.0   \\
      & \textbf{$\mathbf{G}_{\text{{10k}}}^{\text{X}}$}    & 2.392   &  0.099           & 1.803      &  2.909   & 0.0   \\
      & \textbf{$\mathbf{G}_{\text{{500}}}^{\text{Ro}}$}   & 2.983   &  1.091           & 4.668      &  1.195   & 0.0   \\
      & \textbf{$\mathbf{G}_{\text{{10k}}}^{\text{Ro}}$}   & 0.659   &  0.103           & 0.643      &  0.264   & 0.0   \\
      & \textbf{$\mathbf{G}_{\text{{10k}}}^{\text{Unif}}$}  & 1.219   &  0.110           & 0.071      &  0.199   & 0.0   \\
      \parbox[t]{2mm}{\multirow{8}{*}{\rotatebox[origin=c]{90}{\textbf{$T_{\scaleto{\text{MAX}}{3.5pt}} = 50$}}}}  &    &     &   &   &   &     \\
      & \textbf{$\mathbf{G}_{\text{{200}}}^{\text{Exp}}$}  & 4.197   &   2.016          & 0.863      &  0.624   & 0.0   \\
      & \textbf{$\mathbf{G}_{\text{{10k}}}^{\text{Exp}}$} & 1.022   &   0.139          & 0.665      & 0.536    & 0.0    \\
      & \textbf{$\mathbf{G}_{\text{{200}}}^{\text{X}}$}    & 1.206   &   0.139          & 3.600      & 8.967    & 0.0    \\
      & \textbf{$\mathbf{G}_{\text{{10k}}}^{\text{X}}$}    & 2.393   &   0.099          & 0.794      & 2.910    & 0.0   \\
      & \textbf{$\mathbf{G}_{\text{{500}}}^{\text{Ro}}$}  & 2.983   &  1.091           & 3.083      & 0.209    & 0.0   \\
      & \textbf{$\mathbf{G}_{\text{{10k}}}^{\text{Ro}}$}   & 0.659   &  0.103           & 0.076      & 0.264    & 0.0   \\
      & \textbf{$\mathbf{G}_{\text{{10k}}}^{\text{Unif}}$}   & 1.219   &  0.110           & 0.045      & 0.199    & 0.0   \\
      %\midrule
      %\parbox[t]{3mm}{\multirow{6}{*}{\rotatebox[origin=c]{90}{TSP}}} &  &  & &  &  &  &  &  &  &  \\
      %& Fu100                  	&	1.58	&	188	    &	1.09	&	\textit{\underline{0.22}}	&	2.41	&	4.09	&	1.14	&	0.955	&	 \textbf{0.0058} \\
      %& Fu500                  	&	1.58	&	188	    &	1.09	&	\textit{\underline{0.22}}	&	2.41	&	4.09	&	1.14	&	0.955	&	 \textbf{0.0058} \\
      %& Uch100                  	&	1.58	&	188	    &	1.09	&	\textit{\underline{0.22}}	&	2.41	&	4.09	&	1.14	&	0.955	&	 \textbf{0.0058} \\
      %& Uch500                  	&	1.58	&	188	    &	1.09	&	\textit{\underline{0.22}}	&	2.41	&	4.09	&	1.14	&	0.955	&	 \textbf{0.0058} \\
      %\bottomrule
      %OLD AVG!!
      %& AVG                 &       &     &      &       &  \\
      \bottomrule
    \end{tabular} }
  %\captionsetup{type=table, width=.9\linewidth}  %of{table}
  %\end{table}
   %%%%%%%%%%%%%%%%%%%%% tabular 1 %%%%%%%%%%%%%%%%%%%%%
  \end{subtable}%
  \begin{subtable}{.505\linewidth}
  \renewcommand{\arraystretch}{1.3}
%\begin{table}[ht!]
  \centering
  \caption{CVRP}
  \label{table:gap_cvrp}
  \resizebox{\columnwidth}{!}{
  %\scalebox{0.95}{
    %\renewcommand{\arraystretch}{1} % Default value: 1
    % \tiny
    % \footnotesize
    \setlength\tabcolsep{3pt}
    \begin{tabular}{ll|llll|ll}
      \toprule
      & \textbf{Gap (\%)} & \multicolumn{4}{c|}{ML} & \multicolumn{2}{c}{OR}                               \\
      & Dataset                 & POMO\textsubscript{sub}     & BQ\textsubscript{sub}              & SGBS\textsubscript{sub}      & NeuOpt\textsubscript{sub}   &  LKH           & HGS    \\
      \midrule
      \parbox[t]{3mm}{\multirow{8}{*}{\rotatebox[origin=c]{90}{\textbf{$T_{\scaleto{\text{MAX}}{3.5pt}} = 0.7$}}}} & &    &   &    &   &     &     \\
      & \textbf{$\mathbf{G}_{\text{{200}}}^{\text{Exp}}$}  & 1.761   & 2.679            & 2.594      & 27.343 &  2.633   & 0.0    \\
      & \textbf{$\mathbf{G}_{\text{{10k}}}^{\text{Exp}}$}  & 1.774   &  5.506           & 2.638      & 21.231  & 2.999   & 0.0    \\
      & \textbf{$\mathbf{G}_{\text{{200}}}^{\text{X}}$}   & 1.371   & 1.610             & 1.926      & 12.382    & 0.899 & 0.0 \\
      & \textbf{$\mathbf{G}_{\text{{10k}}}^{\text{X}}$}   & \textbf{-2.564}   & 6.153   & \textbf{-2.564}      & 6.085     & 0.580 & 0.0   \\
      & \textbf{$\mathbf{G}_{\text{{500}}}^{\text{Ro}}$}  & 2.239   & 3.548           & 3.098      & 33.845 &  2.555    & 0.0   \\
      & \textbf{$\mathbf{G}_{\text{{10k}}}^{\text{Ro}}$} & 4.468   & 1.958            & 4.123      & 7.669     & 2.150 & 0.0 \\
      & \textbf{$\mathbf{G}_{\text{{10k}}}^{\text{Unif}}$}   & 3.265   & 1.755            & 5.510      & 17.957 &  2.770    & 0.0   \\
      \parbox[t]{2mm}{\multirow{8}{*}{\rotatebox[origin=c]{90}{\textbf{$T_{\scaleto{\text{MAX}}{3.5pt}} = 5$}}}}  &    &     &   &   &   &     \\
      & \textbf{$\mathbf{G}_{\text{{200}}}^{\text{Exp}}$} & 1.858   &  2.084           & 2.525      &  21.928 &  1.813  & 0.0   \\
      & \textbf{$\mathbf{G}_{\text{{10k}}}^{\text{Exp}}$} & 1.471   &  3.012           & 2.278      &  8.965 &  2.005  & 0.0   \\
      & \textbf{$\mathbf{G}_{\text{{200}}}^{\text{X}}$}    & 1.354   &  1.236           & 1.843      & 10.732   & 0.407 & 0.0   \\
      & \textbf{$\mathbf{G}_{\text{{10k}}}^{\text{X}}$}   & \textbf{-3.927}   &  3.155           & \textbf{-2.543}      & 0.358   & 0.382 & 0.0   \\
      & \textbf{$\mathbf{G}_{\text{{500}}}^{\text{Ro}}$}   & 2.342   &  2.711           & 3.074      &  25.961 & 1.904   & 0.0   \\
      & \textbf{$\mathbf{G}_{\text{{10k}}}^{\text{Ro}}$}  & 2.589   &  1.443           & 3.702      &  3.024  & 1.403   & 0.0   \\
      & \textbf{$\mathbf{G}_{\text{{10k}}}^{\text{Unif}}$}   & 3.456   &  1.663           & 5.566      &  7.850 & 2.091  & 0.0   \\
      \parbox[t]{2mm}{\multirow{8}{*}{\rotatebox[origin=c]{90}{\textbf{$T_{\scaleto{\text{MAX}}{3.5pt}} = 50$}}}}  &    &     &   &   &   &     \\
      & \textbf{$\mathbf{G}_{\text{{200}}}^{\text{Exp}}$} & 1.857   &   2.088          & 2.324      &  14.112  & 1.514  & 0.0   \\
      & \textbf{$\mathbf{G}_{\text{{10k}}}^{\text{Exp}}$} & 1.479   &   2.998          & 1.329      & 7.032 &  1.639   & 0.0    \\
      & \textbf{$\mathbf{G}_{\text{{200}}}^{\text{X}}$}    & 1.349   &   1.237          & 1.684      & 8.116    & 0.165 & 0.0    \\
      & \textbf{$\mathbf{G}_{\text{{10k}}}^{\text{X}}$}   & \textbf{-3.935}   &  3.052          & \textbf{-4.224}     & \textbf{-1.966}    & 0.218 & 0.0   \\
      & \textbf{$\mathbf{G}_{\text{{500}}}^{\text{Ro}}$}  & 2.337   &  2.684           & 2.526     & 15.149  & 1.501  & 0.0   \\
      & \textbf{$\mathbf{G}_{\text{{10k}}}^{\text{Ro}}$}  & 2.568   &  1.440           & 1.594      & 1.878    & 1.126 & 0.0   \\
      & \textbf{$\mathbf{G}_{\text{{10k}}}^{\text{Unif}}$}   & 3.439   &  1.681           & 2.989      & 5.402    & 1.690  & 0.0   \\
      %\bottomrule
      %& AVG               &          &           &           &          &                &   \\
      \bottomrule
    \end{tabular}
  }
  %%%%%%%%%%%%%%%%%%%%% tabular 2 %%%%%%%%%%%%%%%%%%%%%
  \end{subtable}
  %\captionsetup{type=table, width=.9\linewidth}
  %\captionsetup{}
%\end{minipage}
\end{table}
%The results on the comparison against LKH demonstrate that the gap to state-of-the-art results from LKH decreased. Notably, we see that some models also outperformed LKH on this task.

%While traditional solvers maintain an edge in average robustness, these findings highlight the potential of ML-based approaches to serve as strong, domain-adaptive solvers, capable of excelling in specific scenarios where classical methods struggle to generalize efficiently.

\section{Discussion and Conclusion}
\label{conclusion}
In this work, we investigated how the performance of classical and neural solvers for routing problems, specifically TSP and CVRP, is influenced by the structure of the input distribution. 
We proposed a simple yet effective subsampling-based training strategy that aligns neural models more closely with specific base node distributions, and benchmarked this approach across a variety of synthetic and real-world-inspired scenarios.
Our experiments show that subsampling can yield competitive and in some cases superior performance, particularly in distribution-specific test settings. Most notably, for CVRP on the Uchoa benchmark, subsampled models outperformed state-of-the-art OR solvers (HGS) under realistic runtime constraints, achieving negative relative gaps.
%, to our knowledge, a first in learned combinatorial optimization.
These results emphasize that distributional alignment plays a critical role in solver performance. While classical meta-heuristics exhibit strong robustness across distributions, they are occasionally outperformed by neural methods trained on structured data. This success suggests that neural solvers are not just scalable, but capable of matching and exceeding handcrafted algorithms when equipped with the right inductive bias, in this case, through data.
However, performance gains are not uniform: TSP continues to favor classical solvers, and neural methods remain sensitive to distributional shift.
Overall, our findings support a central message: solver performance is highly distribution-dependent, and neural methods present a viable, competitive alternative in distribution-matched regimes. This highlights the value of training paradigms that leverage domain-specific structure, rather than pursuing generality alone.
Future research could explore combinatorial problems with more pronounced structure, hybrid strategies that blend full and subsampled training, or adaptive subsampling methods. Scaling these insights to larger or industrial datasets also remains an important direction.

\newpage
\bibliographystyle{plainnat}  % Use plainnat for numeric citations
\bibliography{sample-base}

\newpage 

\appendix

% \section{Technical Appendices and Supplementary Material}
% \hl{Technical appendices with additional results, figures, graphs and proofs may be submitted with the paper submission before the full submission deadline (see above), or as a separate PDF in the ZIP file below before the supplementary material deadline. There is no page limit for the technical appendices.}

\section{Routing Problem Definitions}
\label{app.problem_def}

\paragraph{The TSP.}
The TSP consists of a fully connected graph $G = (V, E)$, where $V$ is a set of $n$ cities, $V = \{v_1, ..., v_n\}$, and $E=\{e_1,\dots,e_{n^2}\}$ is the set of edges. The edges are associated with pairwise distances $d_{ij}, \forall (i,j) \in V$, representing the cost of traveling from customer $i$ to $j$.
The set of feasible solutions $\Omega_G \subseteq \{0,1\}^{n^2}$ is given by the
vectors $x$ for which the corresponding path in $G$ is a Hamiltonian cycle.
Here, $x_i \in \{0,1\}$ displays whether $e_i \in E$ is in the cycle. The
objective function to be minimized is given via
\begin{equation*}
    f_G(x) \coloneqq \sum_{i=1}^{n^{2}} x_id(e_i).
\end{equation*}

\paragraph{The CVRP.}
%$i=1,\ldots, n$
% The CVRP involves a set of customers, $c_i, \phantom{l} i=1,\ldots, n$, a depot node $c_0$ and a cost matrix $d = [d_{ij}]$ of size $(n+1) \times (n+1)$ that reflects the distances between each pair of nodes (including the depot). Furthermore, a demand vector $q_i, \phantom{l} i=1,\ldots, n$ and a vehicle capacity $Q$ characterize the problem first defined in \cite{dantzig1959truck}. A solution, $x$, is given by a set of routes 
% %$r_k, k=1,\ldots, K$ 
% that start and end at the depot. Such a solution is feasible if the sum of customer demands per routes is smaller than or equal to the vehicle capacity and each customer is visited exactly once. The objective to minimize is the sum of total distance consumed by the routes:
%
% \begin{equation}
% \label{cvrp_objective}
%          \hspace{0.1 cm}
%          z = \min \sum_{i,j} d_{ij}x_{ij}
% \end{equation}
%  where $z$ is the solution-, or objective value, i.e. the total distance covered by the routes.
%%%%%     
The CVRP involves a set of customers $C = \{1, \ldots, n\}$, a depot node $0$, all pairwise distances $d_{ij}, \forall (i,j) \in C \cup 0$, representing the cost of traveling from customer $i$ to $j$, a demand per customer $q_i, i \in C$ and a total vehicle capacity $Q$. A feasible solution $x = \{r_1, \ldots, r_{|x|}\}$ is a set of routes $r_i = (r_{i,1}, \ldots, r_{i,N_i})$, which is a sequence of customers, always starting and ending at the depot, such that $r_{i,1}=0$ and $r_{i,N_i}=0$ and $N_i$ denotes the length of the route. It is feasible if the cumulative demand of any route does not exceed the vehicle capacity and each customer is visited exactly once. Let $f(x)$ denote the total distance of a solution $x$, then the goal of the CVRP is to find the feasible solution with minimal cost $x^* = \text{argmin}_x f(x)$.
% For the remainder of this work we will refer to the objective cost of a solution $x$ as $z_x = f(x)$.
%%%%%
%Uchoa et al.:The input consists
%of a set of n + 1 points, a depot and n customers; an (n + 1) × (n + 1) matrix d = [dij ]
%with the distances between every pair of points i and j; an n-dimensional demand vector
%q = [qi] giving the amount to be delivered to customer i; and a vehicle capacity Q. A
%solution is a set of routes, starting and ending at the depot, that visit every customer
%exactly once in such a way that the sum of the demands of the customers of each route
%does not exceed the vehicle capacity. The objective is to find a solution with minimum
%total route distance.

\section{Base Node Distributions}
\label{app:data_generation}

We introduce the underlying distributions (Base Node Distributions) used for subsampling here. 

\paragraph{Uniform Distribution.} The Uniform distribution is the most commonly used distribution to sample coordinates for routing problems in the NCO literature. The two-dimensional coordinates of the customer locations, as well as the CVRP depot location, are sampled uniformly at random from the [0, 1] interval.
For the CVRP, random integers in the interval $[1, 9]$ are sampled and normalize by the capacity for the demands. 
This standard sampling routine was first introduced in \cite{Nazari2018.Reinforcement}.

\paragraph{X Distribution.} Introduced in \citet{uchoa2017new} these realistic data distributions aim at diversifying the pool of existing CVRP problems and introduce additional challenges to benchmark performances. In the realm of NCO, the distribution has been implemented in earlier works \cite{kool2022deep, hottung2019neural} albeit only for the CVRP. We introduce data samplers for the TSP and select the following configurations for the TSP and CVRP base node distributions in this work:
\begin{itemize}
 	\item \textit{Locations:} For the TSP the base node distribution consists of clustered customer coordinates, while for the CVRP the coordinates are sampled, such that half of them are randomly distributed and the other half is clustered (RC). The coordinates are sampled as proposed in \citet{uchoa2017new} on a $[0, 999]^2$ grid and then normalized to the $[0, 1]$ interval.
    We re-implement the sampling routine outlined in \citet{kool2022deep}
    \item \textit{Depot:} For the CVRP the depot node for the base node distribution is selected randomly (R) on the on a $[0, 999]^2$ grid and then normalized.
    \item \textit{Demands:} The CVRP constitutes in our particular base node distribution \emph{unitary demands}. This configuration is coded as the 0th choice in \citet{kool2022deep}, which is why in Table \ref{tab:datasets} the CVRP datasets are denoted to follow the $\text{X}_\textsubscript{RRC0}$.
\end{itemize}

\paragraph{Explosion Distribution.} Customer Coordinates in the Explosion distribution, introduced in \citet{bossek2019evolving}, are first sampled uniformly at random in the $[0, 1]$ interval and mutated afterwards such that coordinates are dispersed around an explosion center (chosen as the (0.5, 0.5) coordinate in this work).
For the explosion CVRP base node distribution, random integers in the interval $[1, 9]$ are sampled and normalize by the capacity for the demands. 

\paragraph{Rotation Distribution.} Similar as in the Explosion distribution, Rotation distributed customer coordinates \cite{bossek2019evolving} are first sampled uniformly at random in the $[0, 1]$ interval and mutated afterwards to mimic a rotation movement in the center of the coordinates, by multiplying coordinate values, depending on their quadrant with a uniformly sampled angle value (in the interval $[1.2, 1.9]$). Similarly to \citet{zhou2023towards}
we sample demands for the Rotation Base Node Distribution as random integers in the $[1, 9]$ interval.

\section{Baselines}
\label{app:baselines}

The baselines selected for the experiments in section \ref{experiments}
are representative of the currently explored methodologies in NCO for routing problems: While POMO \cite{Kwon2020.POMO} and BQ \cite{Drakulic2023.BQNCO} represent established and new constructive methods respectively, SGBS-EAS \cite{Choo2022.Simulationguided} and NeuOpt \cite{Ma2023.Learning} are learned improvement heuristics. Additionally, we introduce the the state-of-the-art OR solvers, LKH3 \cite{helsgaun2017extension} and HGS-CVRP \cite{vidal2022hybrid} here.

\paragraph{POMO}
Policy Optimization with Multiple Optima \cite{Kwon2020.POMO} proposed a training and inference mechanism for constructive models where they adjust the baseline function in the policy gradient, averaging over multiple rollouts with different starting nodes for a problem instance to get a better baseline estimate. For the TSP, this leverages the solution symmetry of the optimal solution, since the optimal solution can be equivalently constructed in multiple ways if solutions are sequentially constructed as any starting node is valid. For the CVRP, one needs to start at the depot and not all nodes afterwards can still guarantee optimality, however \cite{Kwon2020.POMO} show that it still useful for the CVRP in practice. They additionally introduce a related inference mechanism where instead of sampling or doing a beam search, a greedy inference over all possible starting positions is done. The code \footnote{https://github.com/yd-kwon/POMO} is available under the MIT license and is re-implemented for our experiments.
The experimental settings and hyperparameters are kept as in the originally released code. We only adapted the batch size for training to 128 and experimented with a higher embedding dimension (256 instead of 128).

\paragraph{SGBS}
Simulation Guided Beam Search (SGBS) \cite{Choo2022.Simulationguided} is an improved inference strategy for constructive models, which often utilize simple post-hoc tree searches. It evaluates the candidates in the beam more carefully by not only pruning based on the joint probability of the model on the partial solution but also on the actual score based on a rollout. The procedure is further hybridized with efficient active search (EAS) \cite{eas} which fine-tunes the model during the inference procedure. It is noteworthy that the improved pruning scheme and active search also slow down inference and makes SGBS only competitive in longer runtime budget settings. However, the inference scheme works out of the box for trained POMO checkpoints and thus does not need any retraining for the experiments in section \ref{experiments}. Furthermore, we kept the same hyperparameters as in the originally published code for SGBS and selected the default sampling-rollout strategy with 28 iterations EAS iterations.
The code\footnote{https://github.com/yd-kwon/sgbs} is open source under the MIT license.

\paragraph{BQ}
BQ \cite{Drakulic2023.BQNCO} is a constructive model that replaces the encoder-decoder structure used by other models and instead computes the entire network at every construction step with a deep neural network. This allows updating the input state and removing the already decided on nodes, solving only the remaining subproblem. Furthermore, BQ is trained with supervised learning from solver generated solutions. While this makes the model rather costly in terms of training time and resources, the model excels in performs compared to other constructive methods due to its innovative encoding scheme.
For retraining the BQ model on various subsampled distributions, we have mainly kept the hyperparameters suggested in the originally released code. To streamline the training process on smaller compute resources we have reduced the batch size to 128 (rather than 256) and have furthermore used less training samples due to the overhead in computing labels. To this end, we focused training BQ also on the base node distributions rather than the conventionally sampled training instances.
The code\footnote{https://github.com/naver/bq-nco} is open source under the CC BY-NC-SA 4.0 license.

\paragraph{NeuOpt}
NeuOpt \cite{Ma2023.Learning} is a learned improvement method that moves from a current solution to a better one by parametrizing the k-opt local search operator with a neural network. The action space is factorized into a sequence of submoves, called S-, I- and E-Move. An innovative aspect of the method is that it allows for temporary constraint violations, being able to also search through the infeasible space. 
We kept the same hyperparameters as specified in the released code, but downsized the batch size for training. The code is open source\footnote{https://github.com/yining043/NeuOpt} under the MIT license.

 \paragraph{LKH3}
 LKH3 \cite{helsgaun2017extension} is in its core based on an iterative k-opt local search which is enhanced in several ways, such as for example graph sparsification (candidate set generation), and solution recombination to improve solution quality and computational efficiency. It was originally devised for the TSP but now includes extensions to many problem types by minimizing penalty functions to deal with constraints or transforming the problem to an equivalent TSP if such one exists. The source code\footnote{http://webhotel4.ruc.dk/~keld/research/LKH-3/} is available freely for academic and non-commercial use only.

\paragraph{HGS}
HGS is a state-of-the-art meta-heuristic, originally proposed in 2012 by \citet{vidal2012hybrid}. An updated CVRP-specialized version \citet{vidal2022hybrid}, which is available as an open source implementation\footnote{https://github.com/vidalt/HGS-CVRP}, is licensed under the MIT license. HGS is a hybrid strategy of a genetic algorithm and a local search. In brief, it creates a pool of initial solutions and then at each step (i) new solutions are created by recombining solutions from the pool, (ii) these solutions are optimized with a local search, (iii) and the pool is updated with the newly found solutions. For this general strategy to be effective, the pool needs to be precisely controlled with respect to size, solution quality, diversity and a highly effective local search is needed, since this is still the main component for improving solutions. For more details on these strategies, see \cite{vidal2012hybrid, vidal2022hybrid}.

\section{Evaluation Metric}
\label{app:metric}

We evaluate the performances of the trained models with the per-instance computed \textit{Percentage Gap} to the leading state-of-the-art meta-heuristic for the respective problem class, which, in section \ref{experiments}, is LKH for the TSP and HGS-CVRP for the CVRP. 
Thus, for the best obtained solution value $z$ after terminating with a time budget $T_{\scaleto{\text{MAX}}{3.5pt}}$, we compute the relative percentage gap to a leading OR meta-heuristic with solution value $z_{\text{OR}}$ as follows:
\begin{equation}
  \label{fixed_budget_eqn}
  % \hspace{0.1 cm}
  \text{Gap}_{T_{\scaleto{\text{MAX}}{3.5pt}}}(z) = 100 \frac{z - z\textsubscript{OR}}{z_{\text{OR}}}
\end{equation}

We note that, due to time constraints, we evaluated all models once on the respective datasets presented in section \ref{experiments}, which could potentially affect statistical significance. Nevertheless, we argue that generally the re-implemented models exhibit very minor to no variances in evaluation performances for other datasets and tasks.

\end{document}